\documentclass[10pt,twocolumn,letterpaper]{article}

\usepackage[pagenumbers]{cvpr} 

\definecolor{cvprblue}{rgb}{0.21,0.49,0.74}
\usepackage[pagebackref,breaklinks,colorlinks,allcolors=cvprblue]{hyperref}
\usepackage{multirow}
\usepackage{pifont}

\def\paperID{6408} 
\def\confName{CVPR}
\def\confYear{2025}

\title{Towards Satellite Image Road Graph Extraction: A Global-Scale Dataset and A Novel Method}

\makeatletter
\def\thanks#1{\protected@xdef\@thanks{\@thanks
        \protect\footnotetext{#1}}}
\makeatother

\author{Pan Yin\textsuperscript{1*}, Kaiyu Li\textsuperscript{1*}, Xiangyong Cao\textsuperscript{1$\dag$}, Jing Yao\textsuperscript{2}, Lei Liu\textsuperscript{3}, Xueru Bai\textsuperscript{3}, Feng Zhou\textsuperscript{3}, Deyu Meng\textsuperscript{1} \\
\textsuperscript{1}Xi’an Jiaotong University \; \textsuperscript{2}Chinese Academy of Sciences \; \textsuperscript{3}Xidian University\\
{\tt\small yinpan\_22@stu.xjtu.edu.cn, likyoo.ai@gmail.com, caoxiangyong@mail.xjtu.edu.cn} \\
{\tt\small yaojing@aircas.ac.cn, leiliu@xidian.edu.cn, xrbai@xidian.edu.cn} \\
{\tt\small fzhou@mail.xidian.edu.cn, dymeng@mail.xjtu.edu.cn}
}

\begin{document}

\twocolumn[{%
\renewcommand\twocolumn[1][]{#1}%
\maketitle
\vspace{-12mm}
\begin{center}
  \centering
  \captionsetup{type=figure}
  \includegraphics[width=\linewidth]{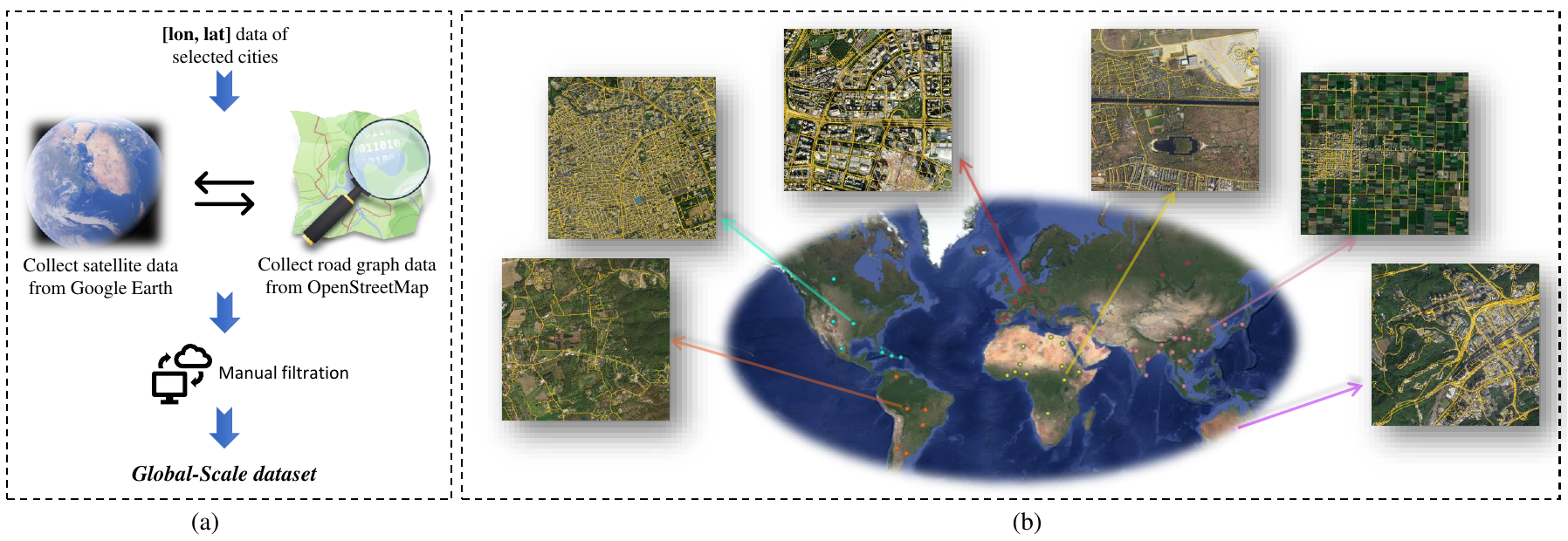}
  \captionof{figure}{(a) The collection pipeline of our \textit{Global-Scale} dataset; (b) The world map showing the region location of the collected images in the \textit{Global-Scale} dataset.}
  \label{fig:dataset}
\end{center}%
}]

\footnotetext{*Equal contribution. \dag Corresponding author.}

\begin{abstract}
Recently, road graph extraction has garnered increasing attention due to its crucial role in autonomous driving, navigation, etc. However, accurately and efficiently extracting road graphs remains a persistent challenge, primarily due to the severe scarcity of labeled data. To address this limitation, we collect a global-scale satellite road graph extraction dataset, \textit{i.e.} Global-Scale dataset. Specifically, the Global-Scale dataset is $\sim20 \times$ larger than the largest existing public road extraction dataset and spans over 13,800 $km^2$ globally. Additionally, we develop a novel road graph extraction model, \textit{i.e.} SAM-Road++, which adopts a node-guided resampling method to alleviate the mismatch issue between training and inference in SAM-Road~\cite{hetang2024segment}, a pioneering state-of-the-art road graph extraction model. Furthermore, we propose a simple yet effective ``extended-line'' strategy in SAM-Road++ to mitigate the occlusion issue on the road. Extensive experiments demonstrate the validity of the collected Global-Scale dataset and the proposed SAM-Road++ method, particularly highlighting its superior predictive power in unseen regions. The dataset and code are available at \href{https://github.com/earth-insights/samroadplus}{https://github.com/earth-insights/samroadplus}.

\end{abstract}    
\section{Introduction}
\label{sec:intro}

In recent years, daily travel has increasingly relied on navigation systems \cite{mohamed2019survey}, particularly with the advent of autonomous driving technology \cite{levinson2011towards}, which has greatly enhanced convenience in everyday life \cite{yurtsever2020survey,nastjuk2020drives}. These advancements demand higher accuracy and real-time performance in extracting road graphs from satellite images \cite{wang2022comprehensive,chen2022road}. Existing approaches to end-to-end road graph extraction can be categorized into two main types: iterative methods \cite{xu2022rngdet, xu2023rngdet++, bastani2018roadtracer} and global-based methods \cite{he2020sat2graph, hetang2024segment}. Iterative methods generate a road graph point by point from the edges of an image. While effective in road graph extraction, this step-by-step connection process can lead to error accumulation and significant computational burdens. In contrast, global-based methods can directly produce a complete road graph, addressing the limitations of iterative approaches. For instance, SAM-Road \cite{hetang2024segment} performs global road graph prediction in two stages: road segmentation and relationship prediction between key points. However, these two stages are uncoupled during training, the first stage employs a conventional road segmentation model to capture the road mask, while the second stage uses node information from the ground truth to train a classifier for node connectivity. This leads to a mismatch during inference, as the classifier's inputs are based on key points selected from the predicted road mask rather than the training data.


To tackle this mismatch issue, we propose a novel ``node sampling" strategy called node-guided resampling. Instead of directly using labeled nodes from the ground truth during classifier training, we resample nodes from the predicted road mask that correspond to the coordinates with the highest probability near the labeled nodes. This approach allows the classifier to leverage training experiences more effectively, resulting in greater consistency between the training and inference processes. Additionally, we recognize that occlusion presents a significant challenge in road graph extraction \cite{yang2024occlusion}. Since models can only extract information from a single overhead view in satellite images, accurately determining the connectivity of road nodes becomes difficult when occlusions (e.g., trees, shadows, etc.) obstruct visibility \cite{dhiman2016continuous,batra2019improved}. To tackle this challenge, we aim to provide the classifier with more contextual information for identifying occluded scenes. Our strategy is based on a straightforward assumption: if road lines exist on either side of two neighboring nodes along a straight road, it is likely that a road also connects these two nodes. Motivated by this assumption, we introduce an ``extended-line" strategy that utilizes the extended line information between two nodes as an additional criterion for determining road connectivity. This enables the model to better navigate occlusion issues in complex environments.

Beyond the algorithms, another bottleneck constraining the road graph extraction task is the limited availability of data. The most commonly used datasets in existing road graph extraction methods, such as City-Scale \cite{he2020sat2graph} and SpaceNet \cite{van2018spacenet}, are constrained by either the number of images or their locations, focusing primarily on urban roads \cite{malin2019accident}. This limited data volume and diversity lead to two main issues: 1) unfaithful evaluations of algorithms and 2) challenges in model generalization capabilities. To address these limitations, we collect a new road graph extraction dataset, \textit{\textbf{Global-Scale}}, which is $\sim20 \times$ larger than existing public datasets. \textit{Global-Scale} encompasses all continents except Antarctica and includes roads from urban, rural, mountainous, and other complex environments. Additionally, to provide a more comprehensive benchmark for algorithm evaluation, we design both in-domain and out-of-domain testing sets within \textit{Global-Scale}, aiming to account for the domain differences present in global road networks. The out-of-domain set consists of data from regions not included in the training set, enhancing the robustness of our evaluations. In summary, the main contributions of this work are threefold:


\begin{itemize}


    \item We establish a novel road graph extraction model, namely SAM-Road++, by coupling the road segmentation and node connectivity prediction sub-networks as a whole. In SAM-Road++, a node-guided resampling strategy is ably introduced to address the mismatch problem between the training and inference phases for the first time.
    

    \item To tackle the issue of road occlusion in satellite images, we propose a novel extended-line strategy that utilizes the correlation between resampled key nodes to facilitate the identification and extraction of connected roads.
    

    
    \item We curate a new benchmark dataset for road extraction, \textit{Global-Scale}, which contains the latest satellite images and faithful road graph maps with larger data volumes, broader coverage, and more diverse scenes, enabling a more comprehensive evaluation of road extraction tasks for the community.    
\end{itemize}






\section{Related Work}
\label{sec:formatting}


\begin{table*}
  \caption{Summary of publicly available road extraction datasets. Grey rows indicate datasets that do not contain graph labels. U, R, and M denote urban, rural, and mountainous areas, respectively.}
  \label{table_data}
  \centering
  \scalebox{0.85}{
  \begin{tabular}{@{}lcccccccccc@{}}
    \toprule[1pt]
    Dataset & Graph Label & Size & Train & Val & Test$^\textit{ID}$ & Test$^\textit{OOD}$ & GSD & Region & Region Type & Time \\
    \midrule[1pt]
    {\color{white!50!black}Massachusetts} \cite{mnih2013machine} & \ding{55} & {\color{white!50!black}$1,500^2$} & {\color{white!50!black}1,108} & {\color{white!50!black}14} & {\color{white!50!black}49} & {\color{white!50!black}\ding{55}} & {\color{white!50!black}1.0} & {\color{white!50!black}Massachusetts} & {\color{white!50!black}U, R} & {\color{white!50!black}2013} \\
    {\color{white!50!black}DeepGlobe} \cite{demir2018deepglobe} & \ding{55} & {\color{white!50!black}$1,024^2$} & {\color{white!50!black}$6,226$} & {\color{white!50!black}243} & {\color{white!50!black}1,101} & {\color{white!50!black}\ding{55}} & {\color{white!50!black}0.5} & {\color{white!50!black}Thailand, Indonesia, India} & {\color{white!50!black}U, R} & {\color{white!50!black}2013} \\
    \midrule[1pt]
    SpaceNet \cite{van2018spacenet} & \checkmark & $400^2$ & 2,167 & \ding{55} & 567 & \ding{55} & 0.3 & Paris, Las Vegas, Shanghai & U & 2018 \\
    City-Scale \cite{he2020sat2graph} & \checkmark & $2,048^2$ & 144 & 9 & 27 & \ding{55} & 1.0 & 20 city in the U.S. & U & 2020 \\
    \midrule[1pt]
    \textit{\textbf{Global-Scale}} & \checkmark & $2,048^2$ & 2,375 & 339 & 624 & 130 & 1.0 & Global & U, R, M & 2024 \\
    \bottomrule[1pt]
  \end{tabular}}
  \vspace{-1em}
\end{table*}

\subsection{Existing Road Extraction Datasets}

Existing road extraction datasets can be divided into two categories: segmentation-labeled datasets \cite{mnih2013machine, demir2018deepglobe,mattyus2015enhancing,zhu2021global} and graph-labeled datasets \cite{van2018spacenet, he2020sat2graph}. 


\noindent \textbf{Segmentation-labeled data} are typical image-mask pairs. The Massachusetts \cite{mnih2013machine} dataset covers a variety of scenes in urban and rural areas, with rich terrain and landform features. The DeepGlobe \cite{demir2018deepglobe} dataset provides more than 10,000 satellite images, covering urban, rural, coastline, and rainforests in Thailand, Indonesia, and India. However, due to the lack of vector information, these datasets are not suitable for road graph extraction tasks.

\noindent \textbf{Graph-labeled data} provides vector graphs of the road network. SpaceNet \cite{van2018spacenet} dataset is first presented in the SpaceNet Challenge\footnote{\url{https://spacenet.ai/challenges/}}. As listed in \cref{table_data}, it contains images that are only 400 $\times$ 400 in size and cover mainly Las Vegas, Paris, and Shanghai. However, the coverage of SpaceNet is limited to urban roads, lacking descriptions of complex terrains like farmland and mountainous regions. On the other hand, the City-Scale \cite{he2020sat2graph} dataset contains only 180 satellite images of 20 cities in the United States, with an image size of 2048 $\times$ 2048. While this dataset offers a broader coverage area than SpaceNet, it still predominantly focuses on urban environments and does not adequately represent non-urban scenes, resulting in an overall insufficient data volume. Consequently, current graph-labeled datasets fall short of meeting modern model requirements \cite{wei2022emergent,kaplan2020scaling} for large-scale and diverse data coverage.

\subsection{Road Graph Extraction}

The existing road graph extraction methods based on satellite images can be divided into two categories: segmentation-based method with post-processing and end-to-end graph-based method.

\noindent \textbf{The segmentation-based method} \cite{gao2018end, li2017road, zhu2021global,chen2023semiroadexnet,lin2020road,abdollahi2020vnet} leverage deep learning technology \cite{lecun2015deep,guo2016deep} to obtain the segmentation mask of the road from images, and then extracts the road graph based on a series of complex post-processing methods \cite{cheng2017automatic,zhang1984fast,li2022hdmapnet}. For instance, Gao et al. \cite{gao2018end} proposed the Multi-Feature Pyramid Network (MFPN), which uses the Feature Pyramid Network (FPN) \cite{lin2017feature} to capture multi-scale semantic features and weighted balanced loss function, improving road extraction accuracy. In contrast, Li et al. \cite{li2017road} developed a CNN-based \cite{kattenborn2021review,li2021survey} framework that extracts road features from small Synthetic Aperture Radar (SAR) image patches \cite{moreira2013tutorial}, identifies candidate road regions, groups, them analyzes road network connectivity with Markov Random Field (MRF). Although segmentation-based methods can generate road graph, their performance in terms of topological connectivity is limited \cite{wang2004image}. Additionally, while they rely on post-processing optimizations, the results remain constrained.

\noindent \textbf{The graph-based methods} are gradually emerging to get a better graph of the road topology in an end-to-end fashion. It can be further divided into iteration-based method \cite{xu2022rngdet, xu2023rngdet++, bastani2018roadtracer} and global-based method \cite{he2020sat2graph, hetang2024segment}. The iteration-based methods build the complete road structure by predicting road nodes (vertices) step by step. The early RoadTracer \cite{bastani2018roadtracer} model started from the initial node and gradually built a road graph with fixed angles and step sizes. 
RNGDet \cite{xu2022rngdet} and RNGDet++ \cite{xu2023rngdet++} combine CNN and Transformer \cite{vaswani2017attention,yuan2021tokens,tarasiou2023vits} to extract features and iteratively predict vertices, which significantly improves the ability to capture the global structure of the road. 
However, due to point-by-point iteration, this type of method is time-consuming, and errors tend to accumulate as points are iterated. In contrast, the global-based method \cite{he2020sat2graph, hetang2024segment} can directly generate a complete road graph with significantly improved efficiency. For example, Sat2Graph \cite{he2020sat2graph} trains the model through graph coding tensors, directly predicts the graph tensor coding of roads, and generates vector graphs through post-processing. Notably, SAM-Road \cite{hetang2024segment} first extracts segmentation masks, and then uses key point location information (from segmentation masks) to directly predict global connectivity. While SAM-Road reduces the number of post-processing steps, its dependence on node labels during training leads to a mismatch between the training and inference phases.



\section{\textit{Global-Scale} Dataset}

\begin{figure*}[t]
  \centering
   \includegraphics[width=1.0\linewidth]{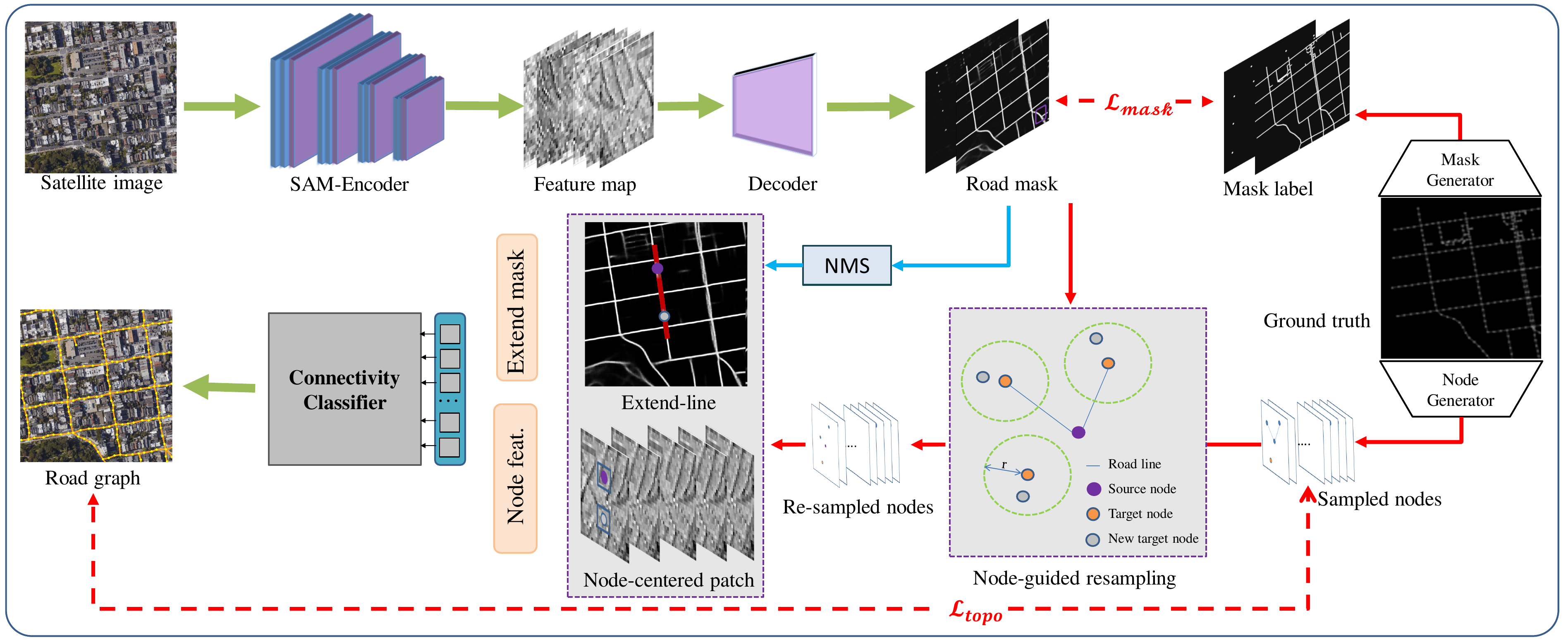}
   \caption{Overview of our proposed SAM-Road++. The \textcolor[RGB]{255, 0, 0}{red} line indicates \textbf{\textit{training only}} and the \textcolor[RGB]{0, 176, 240}{blue} line indicates \textbf{\textit{inference only}}. The satellite image is fed into SAM \cite{kirillov2023segment} Encoder and Decoder to get the feature map and road mask. During training, the proposed node-guided resampling uses the ground truth and road mask to get the re-sampled nodes. In the inference stage, the nodes are obtained using only road masks through Non-Maximum Suppression (NMS). Finally, the connectivity classifier determines whether a road exists between nodes based on the node features and the extended line between the nodes. Both loss functions $\mathcal{L}_{mask}$ and $\mathcal{L}_{topo}$ are binary cross-entropy loss, where $\mathcal{L}_{mask}$ and 
   $\mathcal{L}_{topo}$ are used to supervise the road segmentation and the topological connectivity of the road, respectively.}
   \label{fig:method}
\end{figure*}

A major challenge facing the road graph extraction task is the lack of comprehensive datasets and benchmarks \cite{liu2024review,sun2006road}. Table \ref{table_data} shows the survey results of currently available road extraction datasets, revealing several key issues. First, the existing graph-labeled datasets are small in scale and mainly focus on urban areas. In contrast, segmentation-labeled datasets are not only larger in scale but also cover multiple types such as cities and farmland. Second, current road extraction datasets are limited to the city scale, lacking a global scale road extraction dataset. Finally, although the training set of SpaceNet is large in scale, the size of each image is small (400 $\times$ 400), resulting in a limited coverage area, and roads are usually elongated and connectivity \cite{vznidarivc2011review}, such a size may only include some shorter roads, resulting in limited contextual information. Although City-Scale provides a larger image size (2,048 $\times$ 2,048), which solves the coverage problem, it has a small number of images, especially the test set has only 27 images, which makes the model evaluation susceptible to significant impact from the performance of a single image.

Our \textit{Global-Scale} dataset is a comprehensive road graph resource covering all continents except Antarctica, designed to address the gaps in existing graph-labeled datasets. Figure \ref{fig:dataset} illustrates the collection methodology of the \textit{Global-Scale} dataset. Specifically, We manually selected the longitude and latitude of various types of roads, including urban, rural, and mountainous, using Google Earth \cite{mutanga2019google,lisle2006google}. We then gathered high-quality satellite images from the Google Static Map API \cite{svennerberg2010beginning} based on the selected latitude and longitude information, as well as from the commonly used open-source OpenStreetMap \cite{haklay2008openstreetmap} database to obtain corresponding road graph data as ground truth. Each image has a spatial resolution of 1m/pixel, following the standards set by \cite{he2020sat2graph}. However, the annotation completeness of OpenStreetMap can be affected by variations in population size and economic disparities across regions \cite{herfort2023spatio}. To ensure the annotation completeness of the \textit{Global-Scale} dataset, we conducted a thorough second screening of the annotations after the initial data collection to confirm the integrity of our road graph labels.

Finally, the \textit{Global-Scale} dataset contains 3,468 satellite images, each with a size of 2048 $\times$ 2048, providing extensive spatial coverage and a variety of road types. The dataset is divided into training (2,375 images), validation (339 images), and test sets (624 images). 
Notably, to evaluate the model's inference capability in unseen cities, we additionally collected 130 images from Hong Kong, Shenzhen, and Lucerne, which are not included in the training set, serving as an out-of-domain test set. This design ensures the generalization ability of the model in different geographical environments can be evaluated. In summary, the \textit{Global-Scale} dataset not only provides abundant training resources but also establishes a robust benchmark for future road graph extraction research, allowing researchers to more comprehensively evaluate and compare the performance of different road graph extraction models.



\section{Methodology}

\subsection{Preliminary}


SAM-Road \cite{hetang2024segment} is the first global-based end-to-end road graph extraction method and the first to bring the foundation model \cite{hong2024spectralgpt,osco2023segment} into the field of road graph extraction. Assume that the road graph is $G = (V, E)$, where $V$ is the set of road nodes of the road graph (\textit{i.e.} vertices) and $E$ is the set of road lines (\textit{i.e.} edges). The training phase of SAM-Road consists of two stages. In the first stage, an RGB satellite image is fed into the SAM encoder to extract image feature embeddings. The decoder then predicts per-pixel probabilities for the existence of road lines and nodes in the form of masks. In the second stage, SAM-Road randomly samples nodes (position information) and corresponding edges (road information) from the ground truth.
Using these node pairs, SAM-Road employs the image embeddings of the node pairs obtained in the previous stage as input to train a road classifier to get the probability of a road existing between two nodes.

During the inference phase, ground truth is not available, therefore, SAM-Road selects road nodes from the masks predicted in the first stage as inputs for the classifier of the second stage. As for the selection of nodes, SAM-Road employs a Non-Maximum Suppression (NMS) strategy \cite{hetang2024segment}, which prioritizes nodes with higher prediction probabilities while filtering out surrounding nodes.


\begin{figure}[t]
  \centering
   \includegraphics[width=0.9\linewidth]{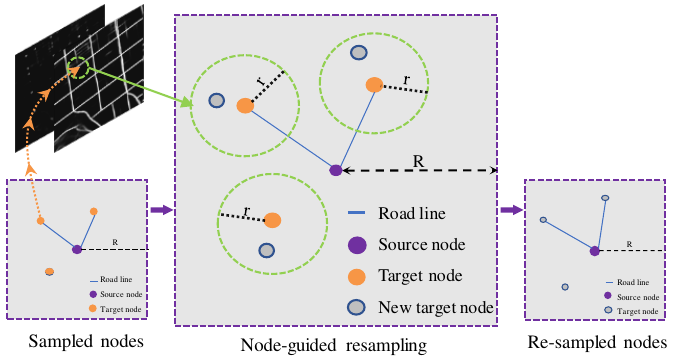}
   \caption{Illustrating the process of node-guided resampling. The sampled nodes are obtained by sampling from the ground truth, and $R$ represents the maximum distance threshold between the source node and target node during the sampling process. Then for each target node, our node-guided resampling strategy will find the maximum probability point of the mask around the target node and save it as the new target node. $r$ represents the maximum distance threshold between the target node and new target node.}
   \label{fig:sampling}
\end{figure}

\begin{figure}[t]
  \centering
   \includegraphics[width=0.9\linewidth]{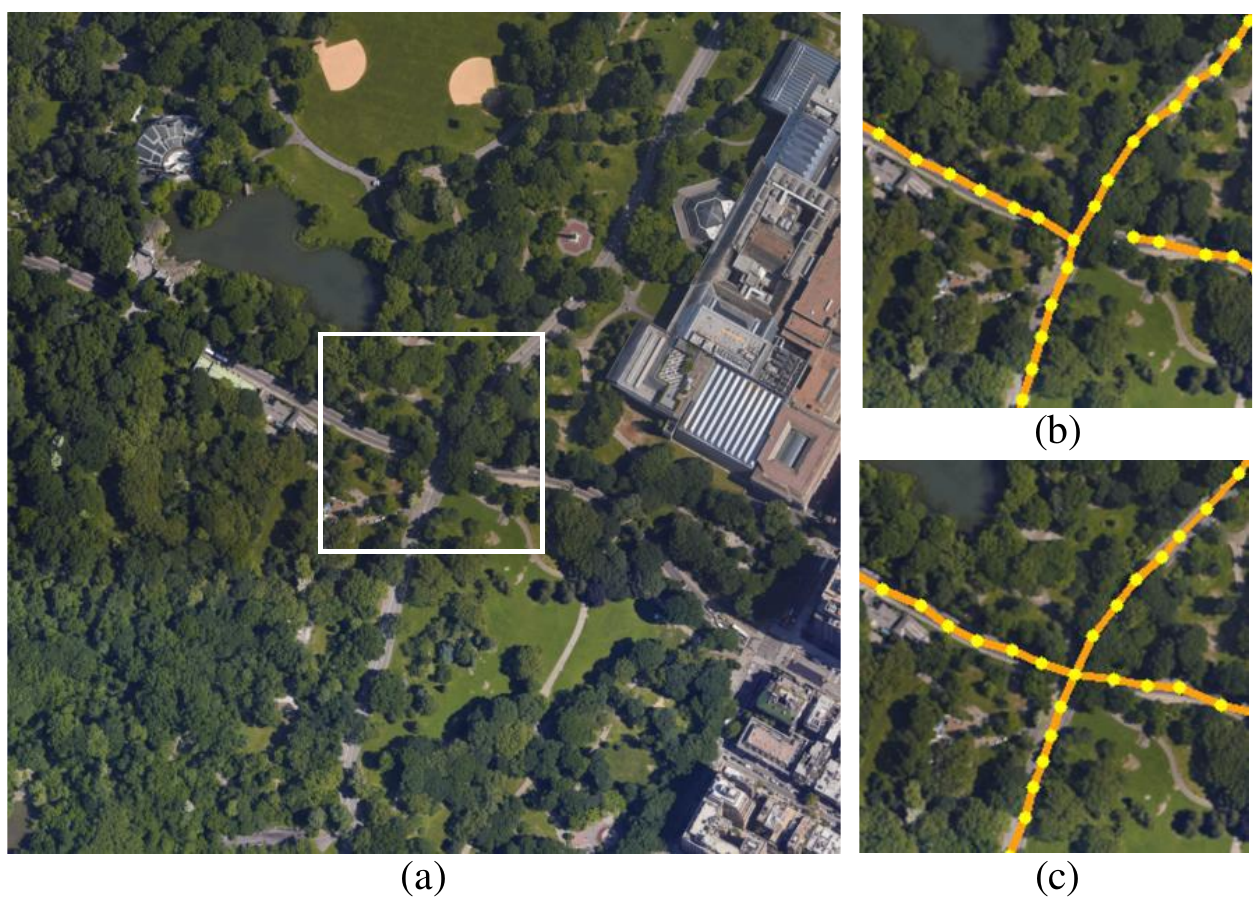}
   \caption{Illustration of occlusion challenge in road nodes connectivity from satellite images. (a) is the raw satellite image, (b) is the prediction of SAM-Road \cite{hetang2024segment}, and (c) is the prediction with the ``extended-line'' strategy.}
   \label{fig:extend}
\end{figure}

However, a significant issue arises from the fact that the classifier trained on ground truth nodes does not perform effectively on the nodes obtained from segmentation masks during the inference process, which indicates a mismatch between training and inference. Consequently, ensuring that the classifier exhibits consistent behavior during both training and inference is critical for road graph extraction. To address this challenge, we propose a novel ``node sampling" strategy.


\subsection{Node-guided resampling}

Our goal is to align the nodes provided to the classifier during training more closely with those used in inference. However, if we directly adopt the inference process for node selection—specifically using the NMS strategy—the connectivity information necessary to supervise the learning of the classifier becomes unavailable. To address this issue, we propose a compromise strategy called node-guided resampling, as illustrated in Figure \ref{fig:sampling}.

First, we sample $N$ nodes (referred to as source nodes) from the ground truth through a node generator. For each source node, we then identify and save all target nodes within a distance $R$, along with the connectivity information between these target nodes and the source node. This process yields the sampled nodes (both source and target). Additionally, to ensure diversity among the sampled nodes, we leverage the inherent degree attributes of the graph for source node selection, specifically favoring nodes with rarer degree attributes, which are easier to sample.


For the sampled nodes, to align more closely with the inference process, we leave the source nodes unchanged and use a radius of 
$r$ to select a new target node with the highest probability, centered around each source node. We then save the coordinate information of this new target node in place of the old one. This results in the creation of re-sampled nodes.
The re-sampled nodes not only retain the source nodes of the ground truth and the connectivity between the nodes, but the position information of the new target nodes also matches the node selection strategy in the inference. 
At the same time, by using the maximum probability of the predicted road mask to pick the new target nodes, it takes full advantage of the experience gained in the previous stage. This strategy compensates for the shortcomings in the training process and effectively improves the consistency of the model during training and inference.

\begin{table*}
  \caption{Comparison of SAM-Road++ with other methods on currently publicly available datasets, The best results are highlighted in bold font, and the second result is underlined. * denotes pre-training on the \textit{Global-Scale} dataset. For all the metrics, larger values indicate better performance.}
  \label{table_compare}
  \centering
  \scalebox{0.9}{
  \begin{tabular}{@{}l|cccc|cccc@{}}
    \toprule[1pt]
    & \multicolumn{4}{c|}{City-Scale} & \multicolumn{4}{c}{SpaceNet}\\
    & F1 & Precision & Recall & APLS & F1 & Precision & Recall & APLS \\
    \midrule[1pt]
    Sat2Graph \cite{he2020sat2graph} & 76.26 & 80.70 & 72.28 & 63.14 & 80.97 & 85.93 & \textbf{76.55} & 64.43 \\
    RNGDet \cite{xu2022rngdet} & 76.87 & 85.97 & 69.87 & 65.75 & 81.13 & 90.91 & 73.25 & 65.61 \\
    RNGDet++ \cite{xu2023rngdet++} & 78.44 & 85.65 & 72.58 & 67.76 & \textbf{82.51} & 91.34 & \underline{75.24} & 67.73 \\
    SAM-Road \cite{hetang2024segment} & 77.23 & \textbf{90.47} & 67.69 & \underline{68.37} & 80.52 & 93.03 & 70.97 & 71.64 \\
    \midrule[1pt]
    Ours & \underline{80.01} & 88.39 & \underline{73.39} & 68.34 & 81.57 & \underline{93.68} & 72.23 & \underline{73.44} \\
    Ours* & \textbf{80.66} & \underline{89.08} & \textbf{74.07} & \textbf{69.55} & \underline{82.07} & \textbf{93.97} & 72.84 &\textbf{74.35} \\
  \bottomrule[1pt]
  \end{tabular}}
\end{table*}

\subsection{``Extended-line'' strategy}

Both in the inference and training process, once the locations of the nodes are determined, the road classifier needs to make road prediction between the source node and target node. However, since the model can only extract information from a single overhead view in satellite images, the determination of connectivity between road nodes is susceptible to occlusion, as shown in Figure \ref{fig:extend}. To tackle this issue, we propose the ``extended-line'' strategy.

For a pair of re-sampled nodes required for connectivity discrimination, we use their coordinates in the image to extract their node-centered patch in feature map obtained from SAM-Encoder, called source and target features. Considering the extensibility of the road and the influence of factors such as tree shadows on road judgment, we believe that the information between the two nodes and the information on their extensions can also effectively assist the model in predicting the existence of the road. Therefore, we uniformly sampled the mask values at both ends of the respective extensions $n$ times by using the road masks previously generated by the model. In addition, we also uniformly sample $m$ times on the line between these two nodes.

\subsection{Inference}

In the inference process, we input the satellite images into the SAM encoder and decoder and obtain the predicted segment masks, including road mask and key point mask. First, we use NMS to deal with the road mask in three steps: 1) we set all pixel values in the mask less than the threshold $t_1$ to 0, and 2) find the coordinates ($x_i$, $y_i$) of the largest probability in the mask and save it to the set $V_1$, then set all pixel values around ($x_i$, $y_i$) within the threshold radius $r_1$ to 0. Next, 3) repeat the second step until all the non-zero pixel points have been traversed. For the key point mask, we use the threshold $l_2$ and the threshold radius $r_2$ to repeat the above three steps to obtain the set $V_2$. Finally, we merge the obtained location sets $V_1$ and $V_2$ together to obtain the final road nodes information, which can be used for the subsequent connectivity prediction. It is worth mentioning that the use of thresholds $l_1$, $l_2$ and radius $r_1$, $r_2$ ensures that we get the node with the largest probability within a certain range, which also corresponds well to the node-guided resampling that we adopt in the training phase.

In the connectivity discrimination stage, each node is considered as a source node, and the nodes in its surrounding range as target nodes. For each pair of source and target nodes, we use the ``extended-line'' strategy to extract information input to the connectivity classifier to predict the probability of whether a road exists between the pair of nodes. Since each node will be a source node and the same roads may be predicted multiple times, we average all the obtained road scores to get the final road graph.

\begin{figure*}[t]
  \centering
   \includegraphics[width=0.85\linewidth]{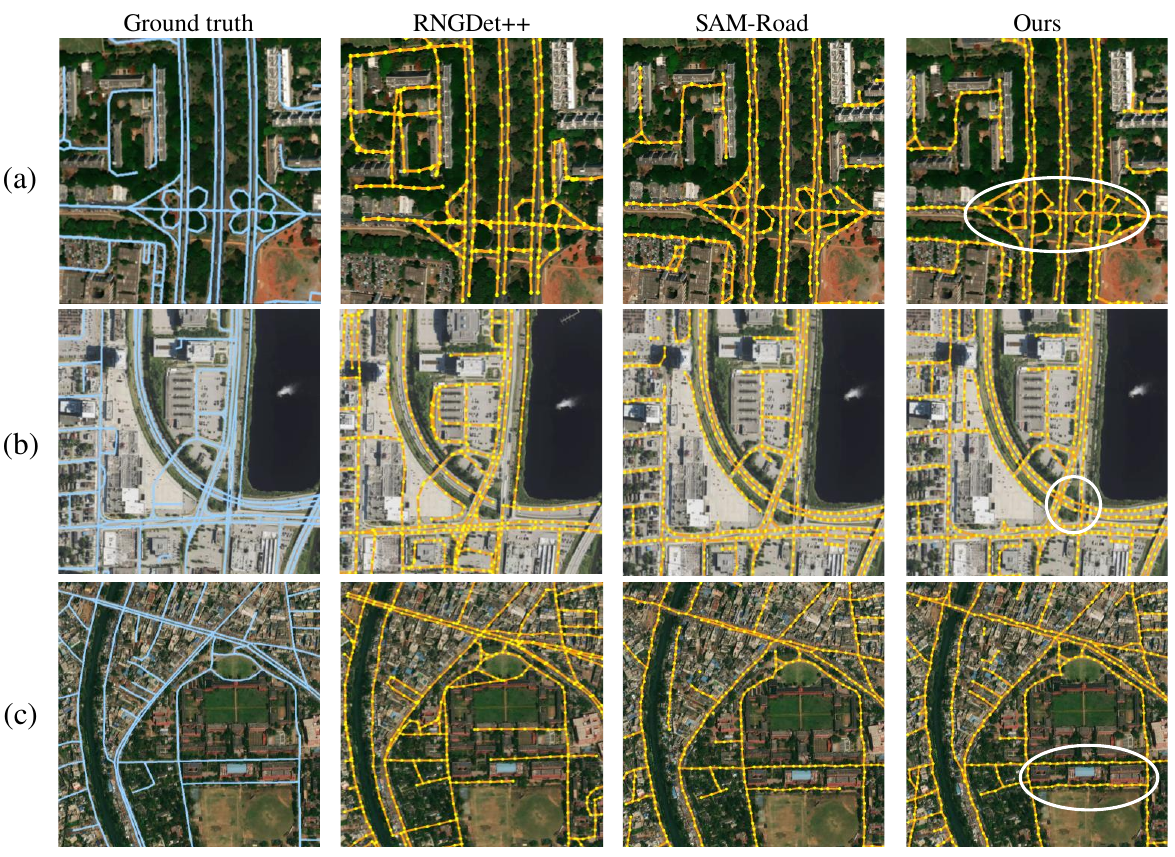}
   \caption{The visualized road network graph predictions of SAM-Road++ and two baseline methods. Better zoom-in and view in color. Overall, the prediction accuracy of SAM-Road++ is higher. In the crossroads regions (a and b),  SAM-Road++ successfully predicted the complexity of multiple roundabouts and overpasses, and in predicting the tree-shaded region c, SAM-Road++'s prediction result is also more complete compared to the other two baselines.}
   \label{fig:vis_compare}
\end{figure*}

\section{Experiments}

\subsection{Datasets}

In this paper, we provide a comprehensive evaluation of three datasets, City-Scale \cite{he2020sat2graph}, SpaceNet \cite{van2018spacenet} and our \textit{Global-Scale}. Table \ref{table_data} shows the basic information of these three datasets, which includes important details such as dataset size and spatial resolution. Additionally, to ensure the fairness of experiments, for SpaceNet, we refer to the pre-processing method of previous works \cite{he2020sat2graph, hetang2024segment}. Specifically, we adjust the spatial resolution of satellite images in the SpaceNet dataset to 1m/pixel.

\subsection{Metrics}

To evaluate the performance of models, we use two metrics: TOPO \cite{biagioni2012inferring} and Average Path Length Similarity (APLS) \cite{van2018spacenet}. 
The TOPO compares the similarity of reachable subgraphs of the same vertices in the predicted graph and ground truth in terms of precision, recall and F1. The APLS metric, on the other hand, focuses on the accuracy of the shortest path between two locations in the road graph. This metric is very sensitive to path quality and is well suited to assessing the accuracy of road connectivity.


\subsection{Implementation}
    For both the City-Scale and \textit{Global-Scale} datasets, we extracted 512 $\times$ 512 pixel image patches from the satellite images with the batch size set to 16, and set the number of sampled source nodes $N$ for each patch to 512. For SpaceNet, the patch size is set to 256 × 256, the batch size to 64, and the number of source nodes $N$ per patch is set to 128.
In the sampling phase, $R$ is set to 16 pixels, and set $r$ to 8 pixels in the sampling phase for all datasets. 
In the implementation of ``extended-line'', since the resolution of all the datasets is 1m/pixel, we set the length on the ends of the respective extended lines to 8 pixels and then set the width of the line to 3 pixels to simulate a road, as well as setting the number of samples $n$ to 15 and $m$ to 20. In the inference phase, we set radius $r_1$ to 16 and radius $r_2$ to 8. The binary classification thresholds $t_1$, $t_2$, and $t_3$ are the thresholds that yield the highest $F_1$ score on the validation set.

We use the Adam optimizer with a basic learning rate of 0.001. For SpaceNet and City-Scale datasets, we train SAM-Road++ until the validation metrics stabilize, following \cite{hetang2024segment}. For \textit{Global-Scale}, SAM-Road++ is trained for 150 epochs.
All experiments are conducted on one 4090 GPU.

\subsection{Comparative Results}

\begin{table*}
  \caption{Comparison with different methods on \textit{global-scale}. Our method significantly outperforms the other methods on the APLS metrics on both in-domain and out-of-domain test sets, and although we do not achieve sota on the precision metric, it performs far better on the more comprehensive F1 metric on both in-domain and out-of-domain test sets.}
  \label{table_benchmark}
  \centering
  \scalebox{0.9}{
  \begin{tabular}{@{}l|cccc|cccc@{}}
    \toprule[1pt]
    & \multicolumn{4}{c|}{\textit{Global-Scale} (In-Domain)} & \multicolumn{4}{c}{\textit{Global-Scale} (Out-of-Domain)}\\
    & F1 & Precision & Recall & APLS & F1 & Precision & Recall & APLS \\
    \midrule[1pt]
    Sat2Graph \cite{he2020sat2graph} & 35.53 & \underline{90.15} & 22.13 & 26.77 & 30.64 & \textbf{84.73} & 19.75 & 22.49 \\
    RNGDet \cite{xu2022rngdet} & 52.59 & 79.89 & 40.72 & 49.43 & 42.62 & 68.79 & 32.60 & 36.33 \\
    RNGDet++ \cite{xu2023rngdet++} & 55.04 & 79.02 & 45.23 & 52.72 & \underline{47.34} & 70.22 & \underline{35.71} & 38.08 \\
    SAM-Road \cite{hetang2024segment} & \underline{59.80} & \textbf{91.93} & \underline{45.64} & \underline{59.08} & 46.64 & \underline{84.54} & 33.81 & \underline{40.51} \\
    Ours & \textbf{62.33} & 88.95 & \textbf{49.27} & \textbf{62.19} & \textbf{48.34} & 82.21 & \textbf{36.04} & \textbf{43.17} \\
  \bottomrule[1pt]
  \end{tabular}}
\end{table*}

\begin{figure}[t]
  \centering
   \includegraphics[width=0.9\linewidth]{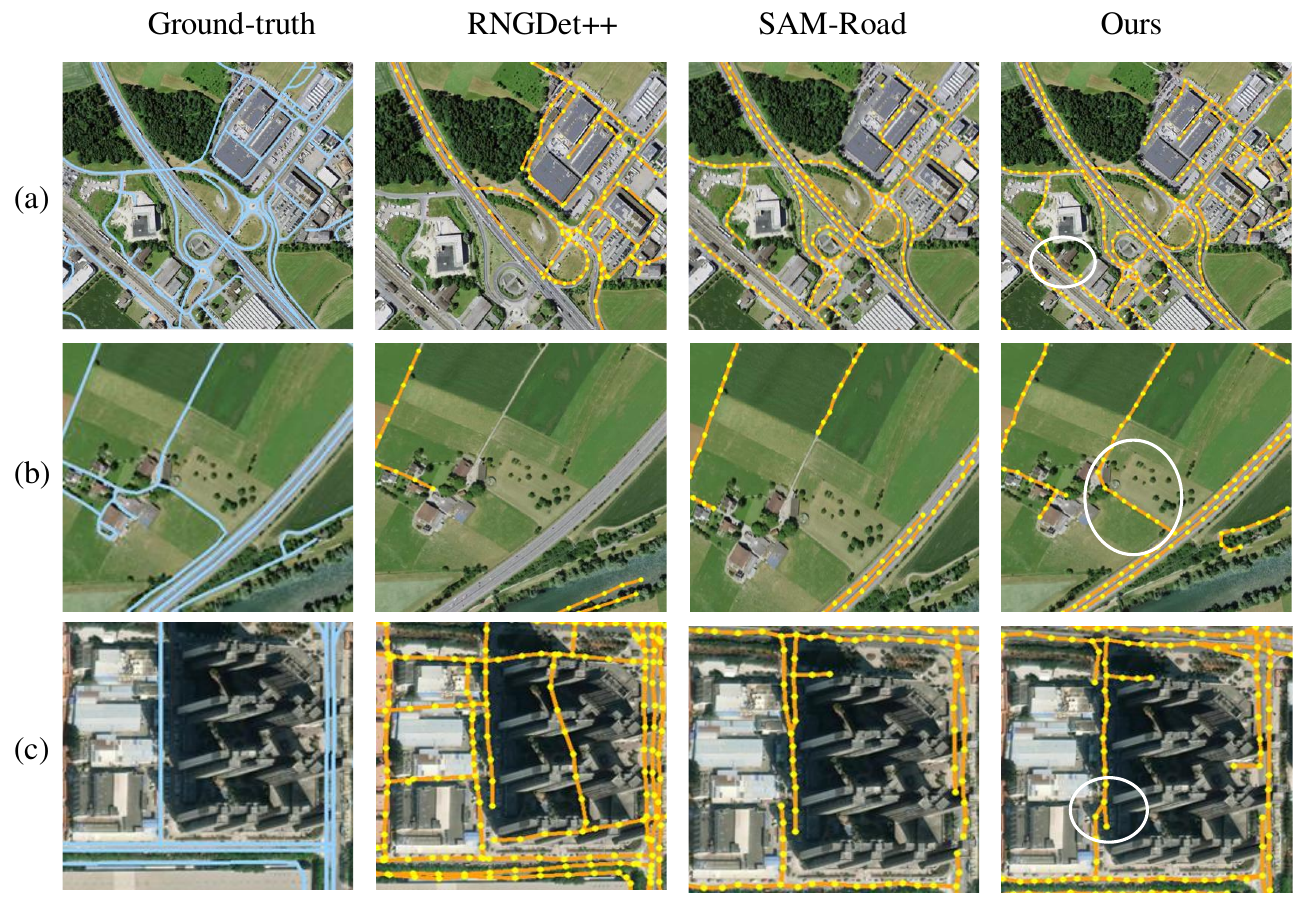}
   \caption{Visualisation of the different methods on the out-of-domain test set. In the rural road region with farmland as context ((a) and (b)), our method successfully avoids road disconnection. In the road region (c) with the shadow of a tall building, our proposed ``extended-line'' strategy accurately predicts the connectivity of the shaded portions.}
   \label{fig:ood}
   \vspace{-1em}
\end{figure}

We evaluate the SAM-Road++ model on the City-Scale and SpaceNet datasets and quantitatively compared it with other methods, as shown in Table \ref{table_compare}. We compare four typical methods, including two iteration-based methods (RNGDet, RNGDet++) and two global-based methods (Sat2Graph, SAM-Road). The experimental results show that our method outperforms existing SOTA methods in F1 metrics for City-Scale and SpaceNet datasets, demonstrating the advantages in overall topology extraction. However, on the City-Scale dataset, although our recall is higher than other methods, the precision is slightly lower than SAM-Road. This is mainly due to the fact that in the inference, a series of node pairs obtained through NMS do not exactly match the ground truth. The node-guided resampling strategy makes the node pairs used in the training phase closer to the inference. As a result, our model is able to identify roads between the node pairs that may deviate from the ground truth, albeit with reduced score.

For the APLS metrics, the performance on the SpaceNet is better than other methods, which indicates that our proposed ``extended-line'' strategy helps the model to infer road lengths closer to the ground truth. Meanwhile, the APLS of our method on the City-Scale dataset is comparable to that of SAM-Road. We note that the City-Scale test set contains so few satellite images (only 27 images) that the per-image prediction has a significant impact on the final APLS metric, which results in the APLS metric being unreliable.
In addition, to further validate the usefulness of our proposed \textit{Global-Scale} dataset, we pre-trained SAM-Road++ on \textit{Global-Scale} and fine-tuned it on the City-Scale and SpaceNet datasets. The experimental results show that the introduction of the \textit{Global-Scale} dataset significantly improves the model's performance in terms of APLS and TOPO metrics on both datasets.


\subsection{Benchmarking on \textbf{\textit{Global-Scale}}}

Figure \ref{fig:vis_compare} and Figure \ref{fig:ood} show the qualitative results of SAM-Road++ on the in-domain and out-of-domain test sets of the global-scale dataset, respectively. As can be seen from the circled areas in regions (a) and (b) in Figure \ref{fig:vis_compare}, when facing complex intersections such as overpasses and multiple roundabouts intersections, the key nodes predicted by the models are not very accurately localized on the road, but owing to the node-guided resampling strategy used in the training phase, SAM-Road++ is able to accurately determine the connectivity of road nodes in this complex scene during the inference phase. The areas circled in Figure \ref{fig:vis_compare}(c) and Figure \ref{fig:ood}(c) are obscured by the shadows of buildings and vegetation on the side of the road, which is an inherent challenge to the road graph extraction task. Previous methods cannot discriminate under such conditions, and the ``extended line'' strategy in SAM-Road++ effectively avoids road breaks based on the extended properties of the road.


For further comparison, we trained the five models mentioned in the previous section on the \textit{Global-Scale} dataset, which contains more comprehensive data, and validated them on both the in-domain test set and out-of-domain test set and the results are shown in Table \ref{table_benchmark}. On the in-domain test set, our method significantly outperforms previous methods in both APLS and TOPO metrics, which indicates that our method outperforms other methods in a wide range of scenarios. Although the overall performance on the out-of-domain test set is lower than that on the in-domain test set, our method still significantly outperforms the other methods, suggesting that our method is more robust to unseen regional data than the other methods, and is more suitable for real-world applications. 

Finally, by comparing the results in Table \ref{table_compare} and Table \ref{table_benchmark}, it can be seen that all five models do not perform as well on \textit{Global-Scale} as they did on City-Scale and SpaceNet, which further demonstrates that \textit{Global-Scale} is a more challenging dataset to better evaluate the generalization ability of models in complex scenarios.

\subsection{Ablation Studies}

\begin{table}
  \caption{Ablation results for ``extended line'' strategy and node-guided resampling.}
  \label{table_ablation}
  \centering
  \scalebox{0.9}{
  \begin{tabular}{@{}cc|cc@{}}
    \toprule[1pt]
    ``extended-line'' & node-guided resampling & APLS & F1 \\
    \midrule[1pt]
    \ding{55} & \ding{55} & 71.64 & 80.52 \\
    \ding{55} & \checkmark & 71.90 & 81.77 \\
    \checkmark & \ding{55} & 73.22 & 80.89 \\
    \checkmark & \checkmark & 73.44 & 81.57 \\
  \bottomrule[1pt]
  \end{tabular}}
  \vspace{-1em}
\end{table}

In this section, we conduct ablation studies to verify the rationality of the design of SAM-Road++, including the node-guided resampling and ``extended-line'' strategy, as shown in Table \ref{table_ablation}. First, we completely remove the node-guided resampling strategy from the training and observe a significant decrease in the model's performance on the F1 metric. This suggests that the node-guided resampling indeed simulates the inference process in training, enabling the model to predict better topologies. Next, we eliminate the "extended-line" strategy and find that the APLS decreases, suggesting that our strategy effectively helps the model predict road lengths that are closer to the ground truth.



\section{Conclusion}



In this paper, we present a large-scale dataset, \textit{Global-Scale}, and a novel method, SAM-Road++, for road graph extraction. The \textit{Global-Scale} encompasses six continents and has been meticulously curated to include a diverse array of scenes, such as urban, rural, and mountainous areas. SAM-Road++ effectively addresses the mismatch between training and inference in global-based methods while mitigating the occlusion challenges inherent in road graph extraction tasks. Extensive experiments demonstrate that \textit{Global-Scale} serves as a more comprehensive and challenging benchmark. In addition, SAM-Road++ achieves superior performance on both existing public datasets and \textit{Global-Scale}, without incurring significant inference costs. Looking ahead, we plan to further expand the \textit{Global-Scale} dataset and endeavor to innovate the paradigm of the road extraction task.


{
    \small
    \bibliographystyle{ieeenat_fullname}
    \bibliography{main}

\begin{thebibliography}{52}
\providecommand{\natexlab}[1]{#1}
\providecommand{\url}[1]{\texttt{#1}}
\expandafter\ifx\csname urlstyle\endcsname\relax
  \providecommand{\doi}[1]{doi: #1}\else
  \providecommand{\doi}{doi: \begingroup \urlstyle{rm}\Url}\fi

\bibitem[Abdollahi et~al.(2020)Abdollahi, Pradhan, and Alamri]{abdollahi2020vnet}
Abolfazl Abdollahi, Biswajeet Pradhan, and Abdullah Alamri.
\newblock Vnet: An end-to-end fully convolutional neural network for road extraction from high-resolution remote sensing data.
\newblock \emph{Ieee Access}, 8:\penalty0 179424--179436, 2020.

\bibitem[Bastani et~al.(2018)Bastani, He, Abbar, Alizadeh, Balakrishnan, Chawla, Madden, and DeWitt]{bastani2018roadtracer}
Favyen Bastani, Songtao He, Sofiane Abbar, Mohammad Alizadeh, Hari Balakrishnan, Sanjay Chawla, Sam Madden, and David DeWitt.
\newblock Roadtracer: Automatic extraction of road networks from aerial images.
\newblock In \emph{Proceedings of the IEEE conference on computer vision and pattern recognition}, pages 4720--4728, 2018.

\bibitem[Batra et~al.(2019)Batra, Singh, Pang, Basu, Jawahar, and Paluri]{batra2019improved}
Anil Batra, Suriya Singh, Guan Pang, Saikat Basu, CV Jawahar, and Manohar Paluri.
\newblock Improved road connectivity by joint learning of orientation and segmentation.
\newblock In \emph{Proceedings of the IEEE/CVF Conference on Computer Vision and Pattern Recognition}, pages 10385--10393, 2019.

\bibitem[Biagioni and Eriksson(2012)]{biagioni2012inferring}
James Biagioni and Jakob Eriksson.
\newblock Inferring road maps from global positioning system traces: Survey and comparative evaluation.
\newblock \emph{Transportation research record}, 2291\penalty0 (1):\penalty0 61--71, 2012.

\bibitem[Chen et~al.(2023)Chen, Li, Wu, Xiong, and Du]{chen2023semiroadexnet}
Hao Chen, Zhenghong Li, Jiangjiang Wu, Wei Xiong, and Chun Du.
\newblock Semiroadexnet: A semi-supervised network for road extraction from remote sensing imagery via adversarial learning.
\newblock \emph{ISPRS Journal of Photogrammetry and Remote Sensing}, 198:\penalty0 169--183, 2023.

\bibitem[Chen et~al.(2022)Chen, Deng, Luo, Li, Junior, Gon{\c{c}}alves, Nurunnabi, Li, Wang, and Li]{chen2022road}
Ziyi Chen, Liai Deng, Yuhua Luo, Dilong Li, Jos{\'e}~Marcato Junior, Wesley~Nunes Gon{\c{c}}alves, Abdul Awal~Md Nurunnabi, Jonathan Li, Cheng Wang, and Deren Li.
\newblock Road extraction in remote sensing data: A survey.
\newblock \emph{International journal of applied earth observation and geoinformation}, 112:\penalty0 102833, 2022.

\bibitem[Cheng et~al.(2017)Cheng, Wang, Xu, Wang, Xiang, and Pan]{cheng2017automatic}
Guangliang Cheng, Ying Wang, Shibiao Xu, Hongzhen Wang, Shiming Xiang, and Chunhong Pan.
\newblock Automatic road detection and centerline extraction via cascaded end-to-end convolutional neural network.
\newblock \emph{IEEE Transactions on Geoscience and Remote Sensing}, 55\penalty0 (6):\penalty0 3322--3337, 2017.

\bibitem[Demir et~al.(2018)Demir, Koperski, Lindenbaum, Pang, Huang, Basu, Hughes, Tuia, and Raskar]{demir2018deepglobe}
Ilke Demir, Krzysztof Koperski, David Lindenbaum, Guan Pang, Jing Huang, Saikat Basu, Forest Hughes, Devis Tuia, and Ramesh Raskar.
\newblock Deepglobe 2018: A challenge to parse the earth through satellite images.
\newblock In \emph{Proceedings of the IEEE conference on computer vision and pattern recognition workshops}, pages 172--181, 2018.

\bibitem[Dhiman et~al.(2016)Dhiman, Tran, Corso, and Chandraker]{dhiman2016continuous}
Vikas Dhiman, Quoc-Huy Tran, Jason~J Corso, and Manmohan Chandraker.
\newblock A continuous occlusion model for road scene understanding.
\newblock In \emph{Proceedings of the IEEE Conference on Computer Vision and Pattern Recognition}, pages 4331--4339, 2016.

\bibitem[Gao et~al.(2018)Gao, Sun, Zhang, Yan, Xu, Sun, Jiao, and Fu]{gao2018end}
Xun Gao, Xian Sun, Yi Zhang, Menglong Yan, Guangluan Xu, Hao Sun, Jiao Jiao, and Kun Fu.
\newblock An end-to-end neural network for road extraction from remote sensing imagery by multiple feature pyramid network.
\newblock \emph{IEEE Access}, 6:\penalty0 39401--39414, 2018.

\bibitem[Guo et~al.(2016)Guo, Liu, Oerlemans, Lao, Wu, and Lew]{guo2016deep}
Yanming Guo, Yu Liu, Ard Oerlemans, Songyang Lao, Song Wu, and Michael~S Lew.
\newblock Deep learning for visual understanding: A review.
\newblock \emph{Neurocomputing}, 187:\penalty0 27--48, 2016.

\bibitem[Haklay and Weber(2008)]{haklay2008openstreetmap}
Mordechai Haklay and Patrick Weber.
\newblock Openstreetmap: User-generated street maps.
\newblock \emph{IEEE Pervasive computing}, 7\penalty0 (4):\penalty0 12--18, 2008.

\bibitem[He et~al.(2020)He, Bastani, Jagwani, Alizadeh, Balakrishnan, Chawla, Elshrif, Madden, and Sadeghi]{he2020sat2graph}
Songtao He, Favyen Bastani, Satvat Jagwani, Mohammad Alizadeh, Hari Balakrishnan, Sanjay Chawla, Mohamed~M Elshrif, Samuel Madden, and Mohammad~Amin Sadeghi.
\newblock Sat2graph: Road graph extraction through graph-tensor encoding.
\newblock In \emph{Computer Vision--ECCV 2020: 16th European Conference, Glasgow, UK, August 23--28, 2020, Proceedings, Part XXIV 16}, pages 51--67. Springer, 2020.

\bibitem[Herfort et~al.(2023)Herfort, Lautenbach, Porto~de Albuquerque, Anderson, and Zipf]{herfort2023spatio}
Benjamin Herfort, Sven Lautenbach, Jo{\~a}o Porto~de Albuquerque, Jennings Anderson, and Alexander Zipf.
\newblock A spatio-temporal analysis investigating completeness and inequalities of global urban building data in openstreetmap.
\newblock \emph{Nature Communications}, 14\penalty0 (1):\penalty0 3985, 2023.

\bibitem[Hetang et~al.(2024)Hetang, Xue, Le, Yue, Wang, and He]{hetang2024segment}
Congrui Hetang, Haoru Xue, Cindy Le, Tianwei Yue, Wenping Wang, and Yihui He.
\newblock Segment anything model for road network graph extraction.
\newblock In \emph{Proceedings of the IEEE/CVF Conference on Computer Vision and Pattern Recognition}, pages 2556--2566, 2024.

\bibitem[Hong et~al.(2024)Hong, Zhang, Li, Li, Li, Yao, Yokoya, Li, Ghamisi, Jia, et~al.]{hong2024spectralgpt}
Danfeng Hong, Bing Zhang, Xuyang Li, Yuxuan Li, Chenyu Li, Jing Yao, Naoto Yokoya, Hao Li, Pedram Ghamisi, Xiuping Jia, et~al.
\newblock Spectralgpt: Spectral remote sensing foundation model.
\newblock \emph{IEEE Transactions on Pattern Analysis and Machine Intelligence}, 2024.

\bibitem[Kaplan et~al.(2020)Kaplan, McCandlish, Henighan, Brown, Chess, Child, Gray, Radford, Wu, and Amodei]{kaplan2020scaling}
Jared Kaplan, Sam McCandlish, Tom Henighan, Tom~B Brown, Benjamin Chess, Rewon Child, Scott Gray, Alec Radford, Jeffrey Wu, and Dario Amodei.
\newblock Scaling laws for neural language models.
\newblock \emph{arXiv preprint arXiv:2001.08361}, 2020.

\bibitem[Kattenborn et~al.(2021)Kattenborn, Leitloff, Schiefer, and Hinz]{kattenborn2021review}
Teja Kattenborn, Jens Leitloff, Felix Schiefer, and Stefan Hinz.
\newblock Review on convolutional neural networks (cnn) in vegetation remote sensing.
\newblock \emph{ISPRS journal of photogrammetry and remote sensing}, 173:\penalty0 24--49, 2021.

\bibitem[Kirillov et~al.(2023)Kirillov, Mintun, Ravi, Mao, Rolland, Gustafson, Xiao, Whitehead, Berg, Lo, et~al.]{kirillov2023segment}
Alexander Kirillov, Eric Mintun, Nikhila Ravi, Hanzi Mao, Chloe Rolland, Laura Gustafson, Tete Xiao, Spencer Whitehead, Alexander~C Berg, Wan-Yen Lo, et~al.
\newblock Segment anything.
\newblock In \emph{Proceedings of the IEEE/CVF International Conference on Computer Vision}, pages 4015--4026, 2023.

\bibitem[LeCun et~al.(2015)LeCun, Bengio, and Hinton]{lecun2015deep}
Yann LeCun, Yoshua Bengio, and Geoffrey Hinton.
\newblock Deep learning.
\newblock \emph{nature}, 521\penalty0 (7553):\penalty0 436--444, 2015.

\bibitem[Levinson et~al.(2011)Levinson, Askeland, Becker, Dolson, Held, Kammel, Kolter, Langer, Pink, Pratt, et~al.]{levinson2011towards}
Jesse Levinson, Jake Askeland, Jan Becker, Jennifer Dolson, David Held, Soeren Kammel, J~Zico Kolter, Dirk Langer, Oliver Pink, Vaughan Pratt, et~al.
\newblock Towards fully autonomous driving: Systems and algorithms.
\newblock In \emph{2011 IEEE intelligent vehicles symposium (IV)}, pages 163--168. IEEE, 2011.

\bibitem[Li et~al.(2022)Li, Wang, Wang, and Zhao]{li2022hdmapnet}
Qi Li, Yue Wang, Yilun Wang, and Hang Zhao.
\newblock Hdmapnet: An online hd map construction and evaluation framework.
\newblock In \emph{2022 International Conference on Robotics and Automation (ICRA)}, pages 4628--4634. IEEE, 2022.

\bibitem[Li et~al.(2017)Li, Zhang, and Wu]{li2017road}
Yue Li, Rong Zhang, and Yunfei Wu.
\newblock Road network extraction in high-resolution sar images based cnn features.
\newblock In \emph{2017 IEEE International Geoscience and Remote Sensing Symposium (IGARSS)}, pages 1664--1667. IEEE, 2017.

\bibitem[Li et~al.(2021)Li, Liu, Yang, Peng, and Zhou]{li2021survey}
Zewen Li, Fan Liu, Wenjie Yang, Shouheng Peng, and Jun Zhou.
\newblock A survey of convolutional neural networks: analysis, applications, and prospects.
\newblock \emph{IEEE transactions on neural networks and learning systems}, 33\penalty0 (12):\penalty0 6999--7019, 2021.

\bibitem[Lin et~al.(2017)Lin, Doll{\'a}r, Girshick, He, Hariharan, and Belongie]{lin2017feature}
Tsung-Yi Lin, Piotr Doll{\'a}r, Ross Girshick, Kaiming He, Bharath Hariharan, and Serge Belongie.
\newblock Feature pyramid networks for object detection.
\newblock In \emph{Proceedings of the IEEE conference on computer vision and pattern recognition}, pages 2117--2125, 2017.

\bibitem[Lin et~al.(2020)Lin, Xu, Wang, Shi, and Chen]{lin2020road}
Yeneng Lin, Dongyun Xu, Nan Wang, Zhou Shi, and Qiuxiao Chen.
\newblock Road extraction from very-high-resolution remote sensing images via a nested se-deeplab model.
\newblock \emph{Remote sensing}, 12\penalty0 (18):\penalty0 2985, 2020.

\bibitem[Lisle(2006)]{lisle2006google}
Richard~J Lisle.
\newblock Google earth: a new geological resource.
\newblock \emph{Geology today}, 22\penalty0 (1):\penalty0 29--32, 2006.

\bibitem[Liu et~al.(2024)Liu, Wu, Lu, Miao, Zhang, Liu, Lu, and Li]{liu2024review}
Ruyi Liu, Junhong Wu, Wenyi Lu, Qiguang Miao, Huan Zhang, Xiangzeng Liu, Zixiang Lu, and Long Li.
\newblock A review of deep learning-based methods for road extraction from high-resolution remote sensing images.
\newblock \emph{Remote Sensing}, 16\penalty0 (12):\penalty0 2056, 2024.

\bibitem[Malin et~al.(2019)Malin, Norros, and Innamaa]{malin2019accident}
Fanny Malin, Ilkka Norros, and Satu Innamaa.
\newblock Accident risk of road and weather conditions on different road types.
\newblock \emph{Accident Analysis \& Prevention}, 122:\penalty0 181--188, 2019.

\bibitem[Mattyus et~al.(2015)Mattyus, Wang, Fidler, and Urtasun]{mattyus2015enhancing}
Gellert Mattyus, Shenlong Wang, Sanja Fidler, and Raquel Urtasun.
\newblock Enhancing road maps by parsing aerial images around the world.
\newblock In \emph{Proceedings of the IEEE international conference on computer vision}, pages 1689--1697, 2015.

\bibitem[Mnih(2013)]{mnih2013machine}
Volodymyr Mnih.
\newblock \emph{Machine learning for aerial image labeling}.
\newblock University of Toronto (Canada), 2013.

\bibitem[Mohamed et~al.(2019)Mohamed, Haghbayan, Westerlund, Heikkonen, Tenhunen, and Plosila]{mohamed2019survey}
Sherif~AS Mohamed, Mohammad-Hashem Haghbayan, Tomi Westerlund, Jukka Heikkonen, Hannu Tenhunen, and Juha Plosila.
\newblock A survey on odometry for autonomous navigation systems.
\newblock \emph{IEEE access}, 7:\penalty0 97466--97486, 2019.

\bibitem[Moreira et~al.(2013)Moreira, Prats-Iraola, Younis, Krieger, Hajnsek, and Papathanassiou]{moreira2013tutorial}
Alberto Moreira, Pau Prats-Iraola, Marwan Younis, Gerhard Krieger, Irena Hajnsek, and Konstantinos~P Papathanassiou.
\newblock A tutorial on synthetic aperture radar.
\newblock \emph{IEEE Geoscience and remote sensing magazine}, 1\penalty0 (1):\penalty0 6--43, 2013.

\bibitem[Mutanga and Kumar(2019)]{mutanga2019google}
Onisimo Mutanga and Lalit Kumar.
\newblock Google earth engine applications, 2019.

\bibitem[Nastjuk et~al.(2020)Nastjuk, Herrenkind, Marrone, Brendel, and Kolbe]{nastjuk2020drives}
Ilja Nastjuk, Bernd Herrenkind, Mauricio Marrone, Alfred~Benedikt Brendel, and Lutz~M Kolbe.
\newblock What drives the acceptance of autonomous driving? an investigation of acceptance factors from an end-user's perspective.
\newblock \emph{Technological Forecasting and Social Change}, 161:\penalty0 120319, 2020.

\bibitem[Osco et~al.(2023)Osco, Wu, de~Lemos, Gon{\c{c}}alves, Ramos, Li, and Junior]{osco2023segment}
Lucas~Prado Osco, Qiusheng Wu, Eduardo~Lopes de Lemos, Wesley~Nunes Gon{\c{c}}alves, Ana Paula~Marques Ramos, Jonathan Li, and Jos{\'e}~Marcato Junior.
\newblock The segment anything model (sam) for remote sensing applications: From zero to one shot.
\newblock \emph{International Journal of Applied Earth Observation and Geoinformation}, 124:\penalty0 103540, 2023.

\bibitem[Sun et~al.(2006)Sun, Bebis, and Miller]{sun2006road}
Zehang Sun, George Bebis, and Ronald Miller.
\newblock On-road vehicle detection: A review.
\newblock \emph{IEEE transactions on pattern analysis and machine intelligence}, 28\penalty0 (5):\penalty0 694--711, 2006.

\bibitem[Svennerberg(2010)]{svennerberg2010beginning}
Gabriel Svennerberg.
\newblock \emph{Beginning google maps API 3}.
\newblock Apress, 2010.

\bibitem[Tarasiou et~al.(2023)Tarasiou, Chavez, and Zafeiriou]{tarasiou2023vits}
Michail Tarasiou, Erik Chavez, and Stefanos Zafeiriou.
\newblock Vits for sits: Vision transformers for satellite image time series.
\newblock In \emph{Proceedings of the IEEE/CVF Conference on Computer Vision and Pattern Recognition}, pages 10418--10428, 2023.

\bibitem[Van~Etten et~al.(2018)Van~Etten, Lindenbaum, and Bacastow]{van2018spacenet}
Adam Van~Etten, Dave Lindenbaum, and Todd~M Bacastow.
\newblock Spacenet: A remote sensing dataset and challenge series.
\newblock \emph{arXiv preprint arXiv:1807.01232}, 2018.

\bibitem[Vaswani(2017)]{vaswani2017attention}
A Vaswani.
\newblock Attention is all you need.
\newblock \emph{Advances in Neural Information Processing Systems}, 2017.

\bibitem[Wang et~al.(2022)Wang, Bayram, and Sertel]{wang2022comprehensive}
Peijuan Wang, Bulent Bayram, and Elif Sertel.
\newblock A comprehensive review on deep learning based remote sensing image super-resolution methods.
\newblock \emph{Earth-Science Reviews}, 232:\penalty0 104110, 2022.

\bibitem[Wang et~al.(2004)Wang, Bovik, Sheikh, and Simoncelli]{wang2004image}
Zhou Wang, Alan~C Bovik, Hamid~R Sheikh, and Eero~P Simoncelli.
\newblock Image quality assessment: from error visibility to structural similarity.
\newblock \emph{IEEE transactions on image processing}, 13\penalty0 (4):\penalty0 600--612, 2004.

\bibitem[Wei et~al.(2022)Wei, Tay, Bommasani, Raffel, Zoph, Borgeaud, Yogatama, Bosma, Zhou, Metzler, et~al.]{wei2022emergent}
Jason Wei, Yi Tay, Rishi Bommasani, Colin Raffel, Barret Zoph, Sebastian Borgeaud, Dani Yogatama, Maarten Bosma, Denny Zhou, Donald Metzler, et~al.
\newblock Emergent abilities of large language models.
\newblock \emph{arXiv preprint arXiv:2206.07682}, 2022.

\bibitem[Xu et~al.(2022)Xu, Liu, Gan, Sun, Wu, Liu, and Wang]{xu2022rngdet}
Zhenhua Xu, Yuxuan Liu, Lu Gan, Yuxiang Sun, Xinyu Wu, Ming Liu, and Lujia Wang.
\newblock Rngdet: Road network graph detection by transformer in aerial images.
\newblock \emph{IEEE Transactions on Geoscience and Remote Sensing}, 60:\penalty0 1--12, 2022.

\bibitem[Xu et~al.(2023)Xu, Liu, Sun, Liu, and Wang]{xu2023rngdet++}
Zhenhua Xu, Yuxuan Liu, Yuxiang Sun, Ming Liu, and Lujia Wang.
\newblock Rngdet++: Road network graph detection by transformer with instance segmentation and multi-scale features enhancement.
\newblock \emph{IEEE Robotics and Automation Letters}, 8\penalty0 (5):\penalty0 2991--2998, 2023.

\bibitem[Yang et~al.(2024)Yang, Zhong, Liu, Lu, and Zhang]{yang2024occlusion}
Ruoyu Yang, Yanfei Zhong, Yinhe Liu, Xiaoyan Lu, and Liangpei Zhang.
\newblock Occlusion-aware road extraction network for high-resolution remote sensing imagery.
\newblock \emph{IEEE Transactions on Geoscience and Remote Sensing}, 2024.

\bibitem[Yuan et~al.(2021)Yuan, Chen, Wang, Yu, Shi, Jiang, Tay, Feng, and Yan]{yuan2021tokens}
Li Yuan, Yunpeng Chen, Tao Wang, Weihao Yu, Yujun Shi, Zi-Hang Jiang, Francis~EH Tay, Jiashi Feng, and Shuicheng Yan.
\newblock Tokens-to-token vit: Training vision transformers from scratch on imagenet.
\newblock In \emph{Proceedings of the IEEE/CVF international conference on computer vision}, pages 558--567, 2021.

\bibitem[Yurtsever et~al.(2020)Yurtsever, Lambert, Carballo, and Takeda]{yurtsever2020survey}
Ekim Yurtsever, Jacob Lambert, Alexander Carballo, and Kazuya Takeda.
\newblock A survey of autonomous driving: Common practices and emerging technologies.
\newblock \emph{IEEE access}, 8:\penalty0 58443--58469, 2020.

\bibitem[Zhang and Suen(1984)]{zhang1984fast}
Tongjie~Y Zhang and Ching~Y. Suen.
\newblock A fast parallel algorithm for thinning digital patterns.
\newblock \emph{Communications of the ACM}, 27\penalty0 (3):\penalty0 236--239, 1984.

\bibitem[Zhu et~al.(2021)Zhu, Zhang, Wang, Zhong, Guan, Lu, Zhang, and Li]{zhu2021global}
Qiqi Zhu, Yanan Zhang, Lizeng Wang, Yanfei Zhong, Qingfeng Guan, Xiaoyan Lu, Liangpei Zhang, and Deren Li.
\newblock A global context-aware and batch-independent network for road extraction from vhr satellite imagery.
\newblock \emph{ISPRS Journal of Photogrammetry and Remote Sensing}, 175:\penalty0 353--365, 2021.

\bibitem[{\v{Z}}nidari{\v{c}} et~al.(2011){\v{Z}}nidari{\v{c}}, Pakrashi, O'Brien, and O'Connor]{vznidarivc2011review}
Ale{\v{s}} {\v{Z}}nidari{\v{c}}, Vikram Pakrashi, Eugene O'Brien, and Alan O'Connor.
\newblock A review of road structure data in six european countries.
\newblock \emph{Proceedings of the Institution of Civil Engineers-Urban design and planning}, 164\penalty0 (4):\penalty0 225--232, 2011.

\end{thebibliography}
}


\end{document}